\documentclass[conference]{IEEEtran}
\usepackage{times}
\usepackage{amsmath}
\usepackage{siunitx}
\usepackage{booktabs}
\usepackage{graphicx}
\usepackage{tikz}
\usetikzlibrary{patterns}
\usepackage{pgfplots}
\pgfplotsset{compat=1.18}
\usepgfplotslibrary{fillbetween}
\usepackage{pgfplotstable}
\usepackage{tabularx}
\usepackage{ragged2e}

\usepackage[numbers]{natbib}
\usepackage{multicol}
\usepackage[bookmarks=true]{hyperref}
\usepackage[acronym]{glossaries}

\usepackage{xcolor}
\definecolor{Black}{HTML}{000000}
\definecolor{Blue}{HTML}{0065bd}
\definecolor{Bluelight}{HTML}{D6E8F7}

\definecolor{Bluestrong}{HTML}{003359}
\definecolor{Red}{HTML}{8C000F}

\definecolor{OrangePP}{HTML}{E97132}
\definecolor{Green}{HTML}{A2AD00}
\definecolor{GreenCR}{HTML}{008000}
\definecolor{LightGray}{HTML}{e7e7e7}
\definecolor{Gray}{HTML}{7f7f7f}
\definecolor{Gray-opac}{HTML}{d8d8d8}
\definecolor{MyDarkBlue}{RGB}{14,40,65}
\definecolor{Red}{RGB}{196,7,27}

    \definecolor{TUMBlue}{HTML}{0065BD}
    \definecolor{TUMBlack}{HTML}{000000}
    \definecolor{TUMWhite}{HTML}{FFFFFF}
    \definecolor{TUMDarkBlue}{HTML}{005293}
    \definecolor{TUMLightBlue}{HTML}{64A0C8}
    \definecolor{TUMLighterBlue}{HTML}{98C6EA}
    \definecolor{TUMGray}{HTML}{999999}
    \definecolor{TUMOrange}{HTML}{E37222}
    \definecolor{TUMGreen}{HTML}{A2AD00}
    \definecolor{TUMLightGray}{HTML}{DAD7CB}

    \definecolor{TUMBlueBrand}{HTML}{3070B3}
    \definecolor{TUMBlueDark}{HTML}{072140}
    \definecolor{TUMBlueDark1}{HTML}{0A2D57}
    \definecolor{TUMBlueDark2}{HTML}{0E396E}
    \definecolor{TUMBlueDark3}{HTML}{114584}
    \definecolor{TUMBlueDark4}{HTML}{14519A}
    \definecolor{TUMBlueDark5}{HTML}{165DB1}
    \definecolor{TUMBlueLight}{HTML}{5E94D4}
    \definecolor{TUMBlueLightDark}{HTML}{9ABCE4}
    \definecolor{TUMBlueLight2}{HTML}{C2D7EF}
    \definecolor{TUMBlueLight3}{HTML}{D7E4F4}
    \definecolor{TUMBlueLight4}{HTML}{E3EEFA}
    \definecolor{TUMBlueLight5}{HTML}{F0F5FA}
    
    \definecolor{TUMYellow}{HTML}{FED702}
    \definecolor{TUMYellowDark}{HTML}{CBAB01}
    \definecolor{TUMYellow1}{HTML}{FEDE34}
    \definecolor{TUMYellow2}{HTML}{FEE667}
    \definecolor{TUMYellow3}{HTML}{FEEE9A}
    \definecolor{TUMYellow4}{HTML}{FEF6CD}
    
    \definecolor{TUMWebOrange}{HTML}{F7811E}
    \definecolor{TUMOrangeDark}{HTML}{D99208}
    \definecolor{TUMOrange1}{HTML}{F9BF4E}
    \definecolor{TUMOrange2}{HTML}{FAD080}
    \definecolor{TUMOrange3}{HTML}{FCE2B0}
    \definecolor{TUMOrange4}{HTML}{FEF4E1}
    
    \definecolor{TUMPink}{HTML}{B55CA5}
    \definecolor{TUMPinkDark}{HTML}{9B468D}
    \definecolor{TUMPink1}{HTML}{C680BB}
    \definecolor{TUMPink2}{HTML}{D6A4CE}
    \definecolor{TUMPink3}{HTML}{E6C7E1}
    \definecolor{TUMPink4}{HTML}{F6EAF4}
    
    \definecolor{TUMBlueBright}{HTML}{8F81EA}
    \definecolor{TUMBlueBrightDark}{HTML}{6955E2}
    \definecolor{TUMBlueBright1}{HTML}{B6ACF1}
    \definecolor{TUMBlueBright2}{HTML}{C9C2F5}
    \definecolor{TUMBlueBright3}{HTML}{DCD8F9}
    \definecolor{TUMBlueBright4}{HTML}{EFEDFC}
    
    \definecolor{TUMRed}{HTML}{EA7237}
    \definecolor{TUMRedDark}{HTML}{D95117}
    \definecolor{TUMRed1}{HTML}{EF9067}
    \definecolor{TUMRed2}{HTML}{F3B295}
    \definecolor{TUMRed3}{HTML}{F6C2AC}
    \definecolor{TUMRed4}{HTML}{FBEADA}
    
    \definecolor{TUMWebGreen}{HTML}{9FBA36}
    \definecolor{TUMGreenDark}{HTML}{7D922A}
    \definecolor{TUMGreen1}{HTML}{B6CE55}
    \definecolor{TUMGreen2}{HTML}{C7D97D}
    \definecolor{TUMGreen3}{HTML}{D8E5A4}
    \definecolor{TUMGreen4}{HTML}{E9F1CB}
    
    \definecolor{TUMGrey1}{HTML}{20252A}
    \definecolor{TUMGrey2}{HTML}{333A41}
    \definecolor{TUMGrey3}{HTML}{475058}
    \definecolor{TUMGrey4}{HTML}{6A757E}
    \definecolor{TUMGrey7}{HTML}{DDE2E6}
    \definecolor{TUMGrey8}{HTML}{EBECEF}
    \definecolor{TUMGrey9}{HTML}{FBF9FA}
    \definecolor{TUMWebWhite}{HTML}{FFFFFF}

\setacronymstyle{long-short}
\defglsentryfmt[\acronymtype]{%
  \ifglsused{\glslabel}%
    {\glsgenacfmt}
    {\glstarget{\glslabel}{\glsgenacfmt}}
}
\newacronym{wm}{WM}{World Model}
\newacronym{vva}{VV\&A}{Verification, Validation \& Accreditation}
\newacronym{sotif}{SOTIF}{Safety of the Intended Functionality}
\newacronym{ad}{AD}{autonomous driving}
\newacronym{ads}{ADS}{Automated Driving System}
\newacronym{fvd}{FVD}{Fr\'echet Video Distance}
\newacronym{cdfvd}{CD-FVD}{content-debiased Fr\'echet Video Distance}
\newacronym{ftd}{FTD}{Fr\'echet Trajectory Distance}
\newacronym{iec}{IEC}{Instruction-Execution Consistency}
\newacronym{ade}{ADE}{Average Displacement Error}
\newacronym{fde}{FDE}{Final Displacement Error}
\newacronym{dtw}{DTW}{Dynamic Time Warping}
\newacronym{sr}{SR}{Success Rate}

\begin{document}

\title{Validate the Dream Before You Trust Its Verdict: Admissibility for World-Model Simulators}


\author{
\authorblockN{
Christian Oefinger\textsuperscript{1},
Finn Rasmus Sch\"afer\textsuperscript{1},
Korbinian Moller\textsuperscript{1},
Mattia Piccinini\textsuperscript{1},
and Johannes Betz\textsuperscript{1}}
\authorblockA{\textsuperscript{1}Autonomous Vehicle Systems Lab\\Technical University of Munich, Garching b. M\"unchen, Germany \\
Email: christian.oefinger@tum.de}}



%

\maketitle

\begin{abstract}
Across robotics, \glspl{wm} are increasingly used to evaluate action policies by simulating the consequences of actions in an imagined world, and returning a success or safety verdict. Yet a verdict is only as trustworthy as the \gls{wm} that produced it, and the \gls{wm} itself needs to be certified. In video-generation \glspl{wm}, fidelity metrics such as \gls{fvd} reward visual realism, but ignore whether the world responds correctly to the policy's actions, including those unseen in training. Classical simulation-based validation assumes a trusted simulator evaluating an untrusted policy, whereas generative \glspl{wm} are themselves unverified learned artifacts. Hence, we argue that any \gls{wm} used as a test oracle must first be accredited before its verdicts can serve as evidence. Building on credibility practices from safety-critical simulation, including \gls{vva}, \gls{sotif}, and scenario-based testing standards, we define an admissibility ladder (L0--L4) that a \gls{wm} must climb before its closed-loop verdicts are accepted as assurance evidence. Our framework is embodiment-agnostic, and is instantiated in \gls{ad}, where assurance methods for traditional simulation are most mature. Applied to two driving \glspl{wm}, the lower rungs reveal a reversal: the model that ranks higher on visual generation quality (L0) ranks lower on action-following (L1--L2), so visual fidelity does not predict the action-robustness a closed-loop verdict depends on.
\end{abstract}

\IEEEpeerreviewmaketitle

\section{Introduction}%
\label{sec:Introduction}%

\Glspl{wm} are generative models that learn an internal representation of an environment's dynamics, letting them predict how the world responds to an agent's actions~\citep{haRecurrentWorldModels2018, hafnerMasteringDiverseControl2025}.
Recent video-generation models turn this prediction into high-fidelity, controllable simulation~\citep{gaoVistaGeneralizableDriving2024, huGAIA1GenerativeWorld2023}, with the potential to reshape how robotic systems are developed and tested.
Across robotics, the role of \glspl{wm} is shifting from \emph{imagining} plausible futures to serving as closed-loop simulators that \emph{test} the action policies that act within them~\citep{houWorldModelRobot2026}:
A \gls{wm} rolls out the consequences of a policy in an imagined world, and returns a verdict on success or safety.
Thus, \glspl{wm} are used as \emph{test oracles} in several robotic embodiments.
\Glspl{wm} serve for instance as (i) a testbed for \textit{manipulation} policies~\citep{quevedoWorldGymWorldModel2025a}, (ii) a benchmark for world-model planning on \textit{legged robots}~\citep{wangTargetBenchCanVideo2025a}, or (iii) a generative simulator for closed-loop evaluation in \textit{autonomous driving}~\citep{yangDriveArenaClosedloopGenerative2024a}.
However, the reliability of \glspl{wm} is typically left unverified, treating their verdicts as \emph{evidence} of real-world behavior, without any basis to assess admissibility.

%

\usetikzlibrary{arrows.meta,calc}

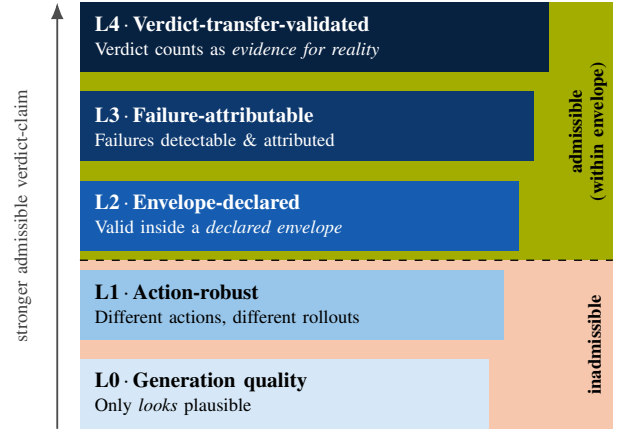
\begin{figure}[t]
\centering
\begin{tikzpicture}[font=\footnotesize, x=1cm, y=1cm]
  \def\bh{0.92}     
  \def\bs{1.18}     
  \def\wL{6.2}      
  \def\dw{0.20}     
  \def\bgr{0.85}    
  \pgfmathsetmacro{\bgW}{\wL+\bgr}
  \pgfmathsetmacro{\split}{(3*\bs+\bh)/2}   
  \pgfmathsetmacro{\topY}{4*\bs+\bh}
  \pgfmathsetmacro{\midOrange}{\split/2}    
  \pgfmathsetmacro{\midGreen}{(\split+\topY)/2}  
  \pgfmathsetmacro{\wzero}{\wL-4*\dw}
  \pgfmathsetmacro{\wone}{\wL-3*\dw}
  \pgfmathsetmacro{\wtwo}{\wL-2*\dw}
  \pgfmathsetmacro{\wthree}{\wL-1*\dw}
  \pgfmathsetmacro{\wfour}{\wL}

  \fill[OrangePP!40] (0,0) rectangle (\bgW,\split);          
  \fill[Green]       (0,\split) rectangle (\bgW,\topY);      

  \fill[Bluelight]    (0,0*\bs) rectangle (\wzero,0*\bs+\bh);
  \node[anchor=west,align=left,inner sep=2pt,text=Black] at (0.12,0*\bs+\bh/2)
     {{\bfseries L0\,$\cdot$\,Generation quality}\\{\scriptsize Only \emph{looks} plausible}};
  \fill[TUMLighterBlue] (0,1*\bs) rectangle (\wone,1*\bs+\bh);
  \node[anchor=west,align=left,inner sep=2pt,text=Black] at (0.12,1*\bs+\bh/2)
     {{\bfseries L1\,$\cdot$\,Action-robust}\\{\scriptsize Different actions, different rollouts}};
  \fill[TUMBlueDark5]  (0,2*\bs) rectangle (\wtwo,2*\bs+\bh);
  \node[anchor=west,align=left,inner sep=2pt,text=white] at (0.12,2*\bs+\bh/2)
     {{\bfseries L2\,$\cdot$\,Envelope-declared}\\{\scriptsize Valid inside a \emph{declared envelope}}};
  \fill[TUMBlueDark2]  (0,3*\bs) rectangle (\wthree,3*\bs+\bh);
  \node[anchor=west,align=left,inner sep=2pt,text=white] at (0.12,3*\bs+\bh/2)
     {{\bfseries L3\,$\cdot$\,Failure-attributable}\\{\scriptsize Failures detectable \& attributed}};
  \fill[TUMBlueDark]   (0,4*\bs) rectangle (\wfour,4*\bs+\bh);
  \node[anchor=west,align=left,inner sep=2pt,text=white] at (0.12,4*\bs+\bh/2)
     {{\bfseries L4\,$\cdot$\,Verdict-transfer-validated}\\{\scriptsize Verdict counts as \emph{evidence for reality}}};

  \draw[dashed,black,line width=0.5pt] (0,\split) -- (\bgW,\split);

  \draw[-{Latex[length=2.2mm,width=1.7mm]},black!70,line width=0.7pt]
     (-0.30,0) -- (-0.30,\topY);
  \node[rotate=90,anchor=south,font=\scriptsize,text=black!75]
     at (-0.52,\topY/2) {stronger admissible verdict-claim};

  \node[rotate=90,font=\scriptsize\bfseries,text=black]
     at (\bgW-0.24,\midOrange) {inadmissible};
  \node[rotate=90,align=center,font=\scriptsize\bfseries,text=black]
     at (\bgW-0.33,\midGreen) {admissible\\(within envelope)};
\end{tikzpicture}
\caption{\textbf{The levels-of-admissibility ladder (L0--L4).}
A generative \gls{wm} used as a closed-loop test oracle earns the right to have its verdict counted as assurance evidence only by climbing the ladder. Verdict \emph{validity} first appears at L2. L0--L1 support no admissibility claim, and all guarantees hold only within the declared operating envelope. Table~\ref{tab:admissibility_ladder} lists the evidence required at each level.}
\label{fig:admissibility_ladder}
\end{figure}

Trusting the verdict assumes a high-fidelity imagined world can substitute for reality.
For video-generation \glspl{wm}, which this paper targets rather than reconstruction-based simulators, visual quality alone does not determine task success~\citep{chenGenerativeEnginesActionable2026, zhangWorldinWorldWorldModels2025}.
Distribution-level video scores, such as the \gls{fvd}~\citep{unterthinerAccurateGenerativeModels2018}, reward perceptual and temporal realism while ignoring physical plausibility and the world's realistic reactions to the policy's actions.
This disconnect has been demonstrated across embodiments.
For legged robots, the best video \glspl{wm} reach only $0.341$ (of a maximum of 1.0) on Target-Bench's overall path-planning score~\citep{wangTargetBenchCanVideo2025a} despite generating visually convincing footage, and an analogous gap holds in \gls{ad}, where open-loop trajectory metrics only partly track closed-loop outcomes~\citep{caoPseudoSimulationAutonomousDriving2025, wangOpenLoopMetricsPredict2026}.
Hence, a world can look right and still judge wrong.

The reason lies in what a verdict implicitly trusts.
Classical simulation-based safety validation relies on trusted simulators, engineered and validated against physics~\citep{stadlerCredibilityAssessmentApproach2022}, and produces verdicts about action policies under test~\citep{beardBlackBoxSafetyValidation2022}.
This relationship is inverted for a generative \gls{wm}.
The simulator, trusted by construction in the classical setting, becomes the component whose reliability is in question, since it is learned and has no ground-truth physics to validate against.
Action policies tested within \glspl{wm} still require validation.
What is new is not that the tested policy needs validation, but that the world judging it must now be validated too, before its verdict can count.
Unlike classical simulators, existing \glspl{wm} carry neither a measurable error nor a certification.

This vulnerability is currently unaddressed.
The established methodologies for licensing safety-critical simulators, \gls{vva}~\citep{balciVerificationValidationAccreditation1997, riedmaierUnifiedFrameworkSurvey2021}, \gls{sotif}~\citep{internationalorganizationforstandardizationisoRoadVehiclesSafety2022}, and scenario-based testing~\citep{RoadVehiclesTest2022}, were all built for classical, physics-based simulators and presuppose an independent ground truth against which the simulator is validated.
None has been adapted to a generative \gls{wm} that is itself the unverified artifact, leaving no principled criterion for when its verdicts may be trusted.

This paper proposes that missing criterion.
We adopt simulation \gls{vva}, \gls{sotif}, and scenario-based testing~\citep{balciVerificationValidationAccreditation1997, RoadVehiclesTest2022, internationalorganizationforstandardizationisoRoadVehiclesSafety2022,  internationalorganizationforstandardizationisoRoadVehiclesTest2023, riedmaierUnifiedFrameworkSurvey2021} methods to form a new \emph{admissibility standard}: criteria, organized as a levels-of-admissibility ladder (L0--L4), that a generative \gls{wm} must meet before its closed-loop verdicts are admissible as evidence~(Figure~\ref{fig:admissibility_ladder}).

The standard is embodiment-agnostic and instantiated in \gls{ad}, where the assurance machinery is most mature and provides the richest source of adaptation.
This paper makes four contributions.
(i) We characterize the \emph{trust inversion} and make precise why a generative \gls{wm}, unlike a classical simulator, is itself the unverified artifact under test, and therefore why existing simulator-validation methods do not license its verdicts.
(ii) We formalize the \emph{action-coverage gap} as an off-policy evaluation problem, identifying the actions outside the training behavior policy as the regime where a \gls{wm}'s verdict is unsupported.
(iii) We construct an \emph{admissibility standard}: a five-level ladder (L0--L4) that, by adapting simulation \gls{vva}, \gls{sotif}, and scenario-based testing to the generative setting, specifies the evidence a \gls{wm} must supply at each level before its closed-loop verdicts count as assurance evidence.
(iv) We instantiate the lower rungs on two driving \glspl{wm} (Vista, Epona), measuring L0, L1, and the L2 horizon, and demonstrate empirically that generation quality and action-robustness decouple (Appendix~\ref{sec:instantiation_ad}).

\section{Related Work}%
\label{sec:Related_Work}%

The use of learned \glspl{wm} to test policies, rather than to plan an agent's actions within them, is a recent but rapidly growing practice.
In general robotics, WorldGym shows that in-simulation policy rankings can correlate with real-world outcomes~\citep{quevedoWorldGymWorldModel2025a}, and World-in-World proposes task success inside imagined rollouts as a general-purpose benchmark signal~\citep{zhangWorldinWorldWorldModels2025}.
In \gls{ad}, DriveArena and Vista already use generative rollouts to score driving behavior~\citep{gaoVistaGeneralizableDriving2024, yangDriveArenaClosedloopGenerative2024a}.
The practice of treating such verdicts as evidence is spreading faster than the principled criteria for trusting them.

Several recent works argue that visual-realism metrics are insufficient to characterize a \gls{wm}'s action-conditioned predictions, pointing to three independent lines of evidence:
(i) Chen and Zhu identify visual conflation and dynamical hallucinations as failure modes that generation scores cannot capture~\citep{chenGenerativeEnginesActionable2026}.
(ii) Action-controllability benchmarks show that models with high visual fidelity can still fail to respond correctly to ego-action inputs~\citep{araiACTBenchActionControllable2024}. (iii) A recent survey frames the gap between generation-quality metrics and closed-loop planning as a central evaluation problem~\citep{zengLatentWorldModels2026}, because fidelity measures the plausibility of the output distribution rather than whether the world reacts correctly to the policy's chosen actions.
We take this growing body of work as a premise rather than a contribution.
Our aim is to specify what must be demonstrated \emph{instead}.

Simulator trustworthiness is addressed in safety-critical engineering through \gls{vva}, which is a structured process for establishing whether a simulator's outputs are sufficiently representative for their intended use~\citep{balciVerificationValidationAccreditation1997}.
In robotics, the reality gap between simulated and real-world dynamics has been formalized and surveyed across robotics domains~\citep{aljalboutRealityGapRobotics2025}.
In \gls{ad}, credibility assessment defines structured evidence requirements for scenario-based virtual testing~\citep{stadlerCredibilityAssessmentApproach2022}, verification, validation, and uncertainty quantification provide structured procedures for bounding predictive uncertainty~\citep{riedmaierUnifiedFrameworkSurvey2021}, and simulator fidelity has been shown to shape which failures a safety-validation process uncovers~\citep{baheriExploringRoleSimulator2023}.
The most recent standards-body treatment flags \glspl{wm} as emerging simulation technology yet defines no admissibility criteria for their verdicts~\citep{SimulationTestingVerification2024}.
Every existing procedure assumes a classical, physics-based simulator. None has been applied to a generative \gls{wm}.

The work most closely premised on our concern is Geng et al.~\citep{gengDeterministicWorldModels2025}, who verify end-to-end controllers against a \emph{deterministic} \gls{wm} via
reachability analysis.
They share the same premise (a \gls{wm}'s outputs must be validated before verdicts count) but take a complementary scope, deterministic and reachability-bounded rather than generative and VV\&A-grounded.
Chen and Zhu~\citep{chenGenerativeEnginesActionable2026} argue that physical grounding is a necessary property of useful simulators. We extend this by specifying when grounding is sufficient for a verdict to count as assurance evidence.
Neither line of work addresses when a generative \gls{wm}'s closed-loop verdict is admissible as assurance evidence.

\section{An Admissibility Standard}%
\label{sec:Admissibility_Standard}%

\subsection{The Credibility Gap}%
\label{subsec:Credibility_Gap}%

Like a generative \gls{wm}, a classical simulator can roll out the consequences of a policy.
Its verdict is credible because its components, such as sensor models or scenario logic, are explicitly engineered and can be validated against physics or recorded data~\citep{ stadlerCredibilityAssessmentApproach2022}.
Additionally, structured procedures such as \gls{vva}~\citep{balciVerificationValidationAccreditation1997}, credibility assessment for scenario-based virtual testing~\citep{stadlerCredibilityAssessmentApproach2022}, and verification, validation, and uncertainty quantification~\citep{riedmaierUnifiedFrameworkSurvey2021} provide principled means of assessing whether their outputs are sufficiently representative for their intended use.
A sim-to-real gap remains, but it is \emph{measurable}: it can be quantified against real-world reference data~\citep{aljalboutRealityGapRobotics2025}, bounded, and reported.
This measurability is the evidence for the verdict's admissibility.
A generative \gls{wm} bypasses this traditional pipeline by learning the simulator directly from data. 
Its sim-to-real gap is immeasurable, since it synthesizes futures for which no recording exists to validate against.
This is the trust inversion of Section~\ref{sec:Introduction}.

A subtler reason a \gls{wm} may mislead its own verdict lies in how it is trained.
Most generative world models learn from logged interaction data, where actions were chosen by the data-collecting behavior policy, such as a human driver, rather than by diverse interventions spanning the action space~\citep{gaoVistaGeneralizableDriving2024, huGAIA1GenerativeWorld2023}.
This yields a model of the observational conditional $p(o' \mid o, a)$, with $o$ and $o'$ the current and next observations and $a$ an action, learned under the behavior policy's action distribution.
This model is reliable only on the $(o,a)$ pairs the behavior policy actually exercised.
A policy under test queries it at the actions \emph{it} selects, including those the behavior policy rarely took or that fall outside the training distribution, where the estimate is subject to \emph{extrapolation error}~\citep{fujimotoOffPolicyDeepReinforcement2019a} and therefore unvalidated.
This is the off-policy evaluation problem~\citep{precupEligibilityTraces2000}, which we call the \emph{action-coverage gap}: the estimates hold only where the target and behavior policies overlap.
Recent evidence confirms it is real.
Action-conditioned reliability depends on training-data diversity~\citep{zhangWorldActionModels2026} and the majority of policy actions are underrepresented in action-labeled training data~\citep{liuWorldActionVerifier2026}.

This problem is not equally severe for all neural simulators. Reconstruction-based simulators that fit a scene representation such as NeRF~\citep{mildenhallNeRFRepresentingScenes2021} or 3DGS~\citep{kerbl3DGaussianSplatting2023} to a recorded capture reduce the sim-to-real gap largely to appearance fidelity, which is partially measurable against the source recording~\citep{lindstromAreNeRFsReady2024}.
The trust problem is qualitatively different for \emph{fully generative} \glspl{wm} that synthesize novel futures from a learned prior.
Their dynamics oracle is the unverified artifact, with no recording to validate against.
This paper targets that class for which no accreditation procedure exists yet.
What remains is to specify what such a \gls{wm} must demonstrate to earn the trust a classical simulator holds by construction.

\subsection{Adapting Existing Certification Frameworks}%
\label{subsec:Adapting_Existing_Frameworks}%
Classical accreditation procedures presuppose a physics-based simulator and therefore do not apply directly to a learned one.
Two of their underlying principles, however, are independent of that assumption, and we adapt them to the generative \gls{wm} setting to derive criteria for when a verdict may be admitted as evidence.
The resulting standard is prescriptive rather than empirical.
It specifies what a \gls{wm} must demonstrate and which evidence it must provide.

The first principle is fidelity-sufficiency. 
\Gls{vva} certifies a simulator not by maximizing realism but by establishing that its fidelity is sufficient for a declared intended use~\citep{balciVerificationValidationAccreditation1997}.
When the intended use is policy evaluation, the link between fidelity and verdict is consequential.
Below a sufficient fidelity, the simulator's outcomes cease to be representative of real-world behavior, so what fails is the evaluation itself, not merely its numerical accuracy~\citep{sagmeisterAnalyzingImpactSimulation2024}.
Adapting this principle to a generative \gls{wm} changes \emph{which} fidelity must be certified.
A \gls{wm} is trained and scored on visual and distributional realism, quantities that are marginal over actions and therefore say nothing about how the world reacts to any particular one~\citep{liuWorldActionVerifier2026, zhangWorldinWorldWorldModels2025}.
A closed-loop verdict depends entirely on those reactions.
We therefore take the object of certification to be the model's \emph{action-conditioned} fidelity: whether, for the actions the policy under test actually selects, the imagined world evolves as it would in the real environment.
Since the \gls{wm} offers no single real-world reference against which this fidelity can be validated at once (Section~\ref{subsec:Credibility_Gap}), sufficiency cannot be certified in one step.
It must be established as a sequence of progressively stronger, separately verifiable claims, which the ladder of Section~\ref{subsec:Admissibility_Ladder} makes explicit.

The second principle is the explicit declaration of operating limits.
\Gls{sotif} and scenario-based testing standards require a system to bound the conditions under which its behavior is claimed~\citep{internationalorganizationforstandardizationisoRoadVehiclesTest2023}, and to treat conditions beyond that boundary as a recognized source of risk~\citep{ internationalorganizationforstandardizationisoRoadVehiclesSafety2022}.
Adapting this principle changes what defines the boundary.
A classical system declares an operating domain specified by design.
A generative \gls{wm} has none, because the region in which its action-conditioned fidelity can hold is fixed by what it was trained on, making the boundary statistical rather than specified.
Establishing where that boundary lies, declaring it, and detecting departures from it is the role of the ladder's envelope level (Section~\ref{subsec:Admissibility_Ladder}), where we make the requirement precise.


\subsection{The Admissibility Ladder}%
\label{subsec:Admissibility_Ladder}%

Each rung of the ladder (Figure~\ref{fig:admissibility_ladder}) licenses a stronger verdict on the policy under test and repurposes an existing diagnostic instrument as an admissibility gate.
No rung may be reached without satisfying the one below.
Table~\ref{tab:admissibility_ladder} lists the evidence required at each level.

\begin{table*}[t]
  \centering
  \caption{\textbf{Required evidence per admissibility level (L0--L4).} Admissibility levels for generative world-model simulators used as closed-loop test oracles.}
  \label{tab:admissibility_ladder}
   \begin{tabularx}{0.9\textwidth}{@{}>{\RaggedRight}p{2.8cm} >{\RaggedRight}X >{\RaggedRight}X@{}}
  \toprule
  \textbf{Level} & \textbf{Admissibility claim} & \textbf{Required evidence} \\
  \midrule
  \textbf{L0}\newline\emph{Generation quality} &
  None: the world only looks plausible and need not respond to actions. &
  Visual and temporal fidelity only~\cite{geContentBiasFrechet2024, luoFVDEnhancedEvaluation2024, unterthinerAccurateGenerativeModels2018}. \\[2pt]\hline

  \textbf{L1}\newline\emph{Action-robust} &
  The world responds to the policy's actions, and semantically different actions yield correspondingly different rollouts. &
  Rollouts vary systematically with the commanded action~\cite{araiACTBenchActionControllable2024, schofieldSimulationBenchmarkingWorld2025}. \\[2pt]\hline

  \textbf{L2}\newline\emph{Envelope-declared} &
  The verdict is valid only within a declared training envelope and rollout horizon, with out-of-distribution detection and refusal outside. &
  Declared training envelope, bounded-drift horizon, and out-of-distribution detection/refusal~\cite{angelopoulosGentleIntroductionConformal2021, chuaDeepReinforcementLearning2018, RoadVehiclesTest2022, internationalorganizationforstandardizationisoRoadVehiclesSafety2022, lakshminarayananSimpleScalablePredictive2017, luoSampleefficientSafetyAssurances2024}. \\[2pt]\hline

  \textbf{L3}\newline\emph{Failure-attributable} &
  Failures at and beyond the boundary are detectable and attributable to simulator vs.\ policy. Out-of-envelope verdicts remain inadmissible. &
  Out-of-distribution failure signatures, plus an attribution protocol separating simulator contributions from policy contributions~\cite{seegertDisengagementAnalysisField2026}. \\[2pt]\hline

  \textbf{L4}\newline\emph{Verdict-transfer-validated} &
  Within the envelope, good-in-sim predicts good-in-reality and the verdict counts as assurance evidence. Outside, L3 keeps failures transparent. &
  Measured in-sim $\leftrightarrow$ real correlation within the envelope, with L3 attribution as the out-of-envelope companion~\cite{quevedoWorldGymWorldModel2025a}. \\
  \bottomrule
  \end{tabularx}

\end{table*}

At \textbf{L0 (Generation quality)}, the simulator is assessed solely on visual and temporal fidelity, typically via Fréchet Video Distance (FVD)~\citep{unterthinerAccurateGenerativeModels2018}.
L0 is a minimum-viability gate.
A model that cannot render a coherent, temporally stable world cannot host a policy test at all, which is why L0 forms the foundation of the admissibility ladder.
Passing this gate is a binary admission, not a quality ranking, and on its own it supports no admissibility claim.
A high fidelity score does not predict the action-following measured at the levels above (Appendix~\ref{subsec:l0_generation_quality} instantiates this rung for two driving \glspl{wm}).
At L0 the simulated world is not required to respond correctly to the policy's actions, the property on which a closed-loop verdict depends.
Content-debiased and successor fidelity measures correct FVD's known biases and estimator flaws~\citep{geContentBiasFrechet2024, luoFVDEnhancedEvaluation2024}, yet they remain L0 instruments.

\textbf{L1 (Action-robust)} requires that the imagined world reacts to the policy's actions, and semantically different actions produce correspondingly different rollouts.
This establishes that the simulator is responsive to the policy under test rather than replaying an action-agnostic future.
Action-controllability benchmarks supply the instruments at this level.
Arai et al.~\citep{araiACTBenchActionControllable2024} measure
whether a rollout follows a commanded action or trajectory, and action-replay analyses probe whether the response degrades under forced trajectory replay~\citep{schofieldSimulationBenchmarkingWorld2025}.
Clearing L1 certifies \emph{differentiation}, not \emph{validity}.
The rollouts vary systematically with the action, but their correspondence to real-world dynamics remains unestablished and is assessed only at higher levels.

\textbf{L2 (Envelope-declared)} adds validity over a bounded region of operation.
Within a declared set of conditions, the simulator's reactions can be checked against real-world behavior, for instance, against measured physical dynamics, and are required to hold.
Outside that region, the simulator's outputs are undefined and do not carry an admissibility claim.
Unlike a classical simulator with an engineered operating domain (in automated driving, an operational design domain~\citep{internationalorganizationforstandardizationisoRoadVehiclesTest2023}), a generative \gls{wm} is bounded by its \emph{training envelope}, the distribution of conditions it was exposed to during training.
The relevant boundary is therefore statistical rather than specified by design.
L2 requires the simulator to declare this envelope and to detect and refuse inputs that fall outside it, mirroring how scenario-based safety and \gls{sotif} bound a system to a declared operating region and treat its violation as a hazard~\citep{internationalorganizationforstandardizationisoRoadVehiclesSafety2022, internationalorganizationforstandardizationisoRoadVehiclesTest2023}.
L2 additionally requires the simulator to bound the rollout horizon over which its reactions remain valid (Appendix~\ref{subsec:l2_horizon_bounded_validity} instantiates the horizon component for two driving \glspl{wm}).
The operative mechanism is out-of-distribution detection, because silently crossing the boundary is what renders a verdict inadmissible.
Viable instruments include conformal prediction, which is distribution-free, gives coverage guarantees under exchangeability, and can be adapted to flag out-of-distribution inputs~\citep{angelopoulosGentleIntroductionConformal2021, luoSampleefficientSafetyAssurances2024}, and ensemble-rollout disagreement, an epistemic-uncertainty heuristic~\citep{chuaDeepReinforcementLearning2018, lakshminarayananSimpleScalablePredictive2017}.

\textbf{L3 (Failure-attributable)} concerns behavior at and beyond the envelope boundary.
It requires that failures be detectable and attributable.
The framework must distinguish failures induced by the simulator from failures of the policy under test.
Verdicts outside the declared envelope remain inadmissible, but at L3 the failures that occur there are transparent rather than silent.
The required evidence is a set of out-of-distribution failure signatures together with an attribution protocol that separates simulator contributions from policy contributions at the boundary.
The supporting instrument is calibration against real failure data, such as field disengagement and criticality records~\citep{seegertDisengagementAnalysisField2026}.

\textbf{L4 (Verdict-transfer-validated)} is the highest rung and the only one resting on \emph{external}, empirical rather than internal evidence.
Inside the declared envelope, it requires a demonstrated correlation between in-simulation and real-world outcomes, so that good in simulation predicts good in reality, and the verdict counts as assurance evidence.
Outside the envelope, L3's attribution capability ensures that failures remain transparent rather than hidden.
The required evidence is a measured correlation between in-simulation and real outcomes within the envelope, complemented outside it by L3's failure attribution.
The instrument is a real-to-simulation policy-performance correlation, as reported for manipulation by WorldGym~\citep{quevedoWorldGymWorldModel2025a}.

\section{Discussion and Conclusion}%
\label{sec:conclusion}%

As generative \glspl{wm} are adopted to test policies, the robotics community increasingly needs to agree on when a \gls{wm}'s verdict can be trusted as evidence.
The admissibility ladder is our proposal for structuring that decision.
Rather than a mandatory standard, the ladder offers a structured vocabulary for that conversation.
By naming the levels a \gls{wm} can demonstrably clear, and the evidence required at each, the ladder makes the implicit assumptions behind in-simulation verdicts explicit and auditable.
Concretely, we suggest that developers proposing generative \glspl{wm} as closed-loop test oracles support their fidelity scores and task metrics with a credibility argument: which rungs does the system clear, and on what evidence?

Three directions remain open.
First, the levels are derived from first principles and by analogy with classical simulation validation.
Their application to existing generative \glspl{wm} is still in progress, as is the test of whether the levels (L0--L4) are jointly sufficient, mutually exclusive, and correctly ordered.
Appendix~\ref{sec:instantiation_ad} takes a first step in this direction, instantiating L0, L1, and the horizon component of L2 on two driving \glspl{wm}.
Second, embodiment-agnostic L2 protocols that define a training envelope and reliably detect out-of-distribution inputs do not yet exist.
Developing and benchmarking such detectors across manipulation, locomotion, and navigation is required to make L2 operational.
Third, L3 attribution requires real-world failure data, which is scarce and domain-specific.
Building calibrated failure datasets together with an attribution protocol that separates simulator from policy contributions is the concrete prerequisite for L3.
We present the ladder as an initial proposal, to be refined as the community applies it across embodiments and tasks.

\clearpage


\bibliographystyle{plainnat}
\bibliography{references}

\clearpage
\appendix

\section{Instantiating the Admissibility Ladder in Autonomous Driving}
\label{sec:instantiation_ad}
This appendix takes a first step toward applying the admissibility ladder to existing generative \gls{wm}.
We instantiate the lower rungs of the ladder (L0--L2) as concrete, reproducible measurements and apply them to two open video-generation \glspl{wm}, placing each on the highest rung its measured evidence supports.
The intent is illustrative rather than a benchmark: a worked example showing that the proposed levels can be operationalized and that they discriminate between models, not a ranking of systems or a general-purpose evaluation protocol.

\subsection{Experimental Scope}
\label{subsec:experimental_scope}

Instantiating the lower rungs in \gls{ad} requires generative \glspl{wm} that can be both run and driven.
The model must be open-weight, so that its rollouts can be regenerated and scored.
Furthermore, it must be action-conditioned, since only a model steered by an external control input can serve as a closed-loop oracle for the policy under test.
Two driving \glspl{wm} meet these criteria: Vista~\citep{gaoVistaGeneralizableDriving2024} and Epona~\citep{zhangEponaAutoregressiveDiffusion2025}.
Both accept an injected ego trajectory and differ in generation paradigm: Vista builds on clip-level video diffusion, and Epona on autoregressive diffusion that predicts one frame at a time.
We evaluate both on the nuScenes-based ACT-Bench split~\citep{araiACTBenchActionControllable2024}, which supplies the action templates
and the pretrained action-recognition instrument that the lower rungs rely on.

We use two models, chosen for three reasons.
First, they come from successive years (Vista from 2024 and Epona from 2025), so a difference in placement reflects progress between them.
Second, Vista has already been evaluated by ACT-Bench~\citep{araiACTBenchActionControllable2024}, which provides an external score (\qty{30.7}{\percent}) that we can use to validate our pipeline, whereas Epona has not been evaluated, so its placement is a new result.
Third, and most importantly, showing that the rungs are independent rather than redundant requires at least two models whose rankings can disagree.

\subsection{Measurement Protocol}
\label{subsec:measurement_protocol}

We operationalize each rung as one or more reproducible metrics, each paired with an explicit decision rule that maps a measured value to a placement. 
In keeping with the ladder, we do not introduce new instruments but repurpose existing ones, and a model is placed at the highest rung whose evidence clears its decision rule.
L0 draws on video-distribution metrics. L1 and L2 instead share a single instrument from ACT-Bench~\citep{araiACTBenchActionControllable2024}: the ACT-Estimator, a pretrained model that infers which maneuver a generated rollout executed and recovers its trajectory.
We reuse this estimator rather than build our own recovery method, since ACT-Bench already trains and validates it, which keeps the L1 and L2 measurements
independent of our implementation.
All metrics are computed on a fixed stratified subsample of the ACT-Bench split: 400 nuScenes clips drawn evenly across ACT-Bench's eight commandable maneuver categories, each paired with its commanded action template.
Checkpoints, seeds, and thresholds are pinned for reproducibility.

\subsubsection{L0: Generation Quality}
\label{subsec:l0_generation_quality}

L0 measures the visual and temporal fidelity of the generated video against real driving footage, the property that existing fidelity metrics target.
We report three Fr\'echet distances between generated and held-out real nuScenes clips.
The
\gls{fvd}~\citep{unterthinerAccurateGenerativeModels2018} compares I3D video features and is the field standard.
The \gls{cdfvd}~\citep{geContentBiasFrechet2024} replaces these with VideoMAE-v2 features~\citep{wangVideoMAEV2Scaling2023}, which respond more to temporal and motion artifacts
than to per-frame appearance.
The \gls{ftd}~\citep{zhouDrivingGenComprehensiveBenchmark2026a} instead compares the distribution of generated ego trajectories, embedded with a pretrained motion encoder, and so captures trajectory realism rather than pixel realism.
Because Fr\'echet distances depend on sample size, all three are computed at a matched number of clips, so that the two models are directly comparable.
L0 carries no admissibility claim: any functioning generator clears it, and we report it only as the foil against which the higher rungs are read.

Both models clear L0 (Table~\ref{tab:results_table}). Vista attains the better pixel- and motion-level scores (\gls{fvd} 151.3 and \gls{cdfvd} 51.6, against 159.4 and 86.1 for Epona), whereas Epona attains the better trajectory-distribution score (\gls{ftd} 2.59 against 2.72).
Visual fidelity therefore already separates from trajectory fidelity within L0, a split we return to in Section~\ref{subsec:decoupling}.

\begin{table}[htbp]
  \centering
  \caption{\textbf{Measured ladder metrics for Vista and Epona (L0--L2).} Performance comparison of Vista and Epona across the admissibility-ladder metrics. $h^*$ is the maximum admissible rollout horizon, the largest evaluated horizon at which ADE stays within the $1.8$\,m band (half a $3.6$\,m US lane~\citep{americanassociationofstatehighwayandtransportationofficialsPolicyGeometricDesign2018}; see Fig.~\ref{fig:horizon_drift}).}
  \label{tab:results_table}
  \begin{tabularx}{\columnwidth}{XXXX}
\toprule
\textbf{Rung} & \textbf{Metric} & \textbf{Vista (2024)} & \textbf{Epona (2025)} \\
\midrule
L0 & FVD $\downarrow$ & \textbf{151.3} & 159.4 \\
 & CD-FVD $\downarrow$ & \textbf{51.6} & 86.1 \\
 & FTD $\downarrow$ & 2.72 & \textbf{2.59} \\
\midrule
L1 & IEC $\uparrow$ & 0.33 & \textbf{0.54} \\
 & ADE $\downarrow$ (m) & 4.56 & \textbf{2.35} \\
 & DTW $\downarrow$ & 97.3 & \textbf{41.5} \\
 & Success $\uparrow$ & 0.08 & \textbf{0.28} \\
\midrule
L2 & $h^*$ $\uparrow$ (s) & 1.6 & \textbf{3.2} \\
 & drift $\downarrow$ (m/s) & 1.06 & \textbf{0.53} \\
\bottomrule
\end{tabularx}
\end{table}

\subsubsection{L1: Action-Robustness}
\label{subsec:l1_action_robustness}

L1 asks whether a rollout follows the action it was commanded, rather than replaying a future that ignores the policy.
For each clip, we feed the commanded action template to the model, generate a rollout, and apply the ACT-Estimator to recover the executed maneuver and trajectory.
The recovered values are then compared with the command by four metrics.
\Gls{iec}~\citep{araiACTBenchActionControllable2024} is the fraction of rollouts whose executed maneuver matches the commanded one across the eight commandable maneuver categories,\footnote{ACT-Bench's estimator classifies nine maneuvers, but only eight are commandable: no action template instructs the ego to remain \emph{stopped}, so it can never appear as a commanded maneuver. IEC is scored over these eight (chance $1/8$); the estimator-reconstruction check below retains all nine to match the estimator's classification task.} where chance is $1/8$.
\Gls{ade} and \gls{fde} are the mean and final Euclidean distance between the commanded and recovered trajectories~\citep{wilsonArgoverse2Next2021}.
\Gls{dtw} measures the same mismatch under an elastic time alignment, so that a correctly shaped but delayed maneuver is not over-penalized~\citep{sakoeDynamicProgrammingAlgorithm1978}.
\Gls{sr} is the fraction of rollouts whose \gls{fde} falls below \qty{3}{\meter}, the default threshold of the \texttt{success\_rate} metric in DrivingGen's~\citep{zhouDrivingGenComprehensiveBenchmark2026a} reference implementation. This mirrors the complementary miss-rate convention of motion-forecasting benchmarks, which threshold the endpoint error at a fixed distance~\citep{wilsonArgoverse2Next2021} (\qty{2}{\meter} in Argoverse).
We adopt it rather than define our own.
A model clears L1 if its \gls{iec} lies significantly above chance, which shows the rollout tracks the commanded maneuver rather than ignoring it.
Clearing L1 certifies responsiveness, not that the resulting dynamics are physically correct, which the higher rungs address.

We validate the instrument before trusting any generated-video score.
On real nuScenes clips the ACT-Estimator recovers trajectories to a mean \gls{ade} of \qty{0.77}{\meter}, and a reconstruction of ACT-Bench's rule-based maneuver classifier reproduces the estimator's maneuver predictions on \qty{93.4}{\percent} of the clips it labels, close to the \qty{94.0}{\percent} ACT-Bench reports.
Scoring Vista's ACT-Bench-released rollouts, our pipeline reproduces the published score (\gls{iec} \qty{30.7}{\percent}).
We then apply the identical scoring to Epona rollouts that we generate with an adapter mapping each template to Epona's action format, as ACT-Bench did not evaluate Epona.
Because Epona's action format includes a heading the templates lack, the adapter synthesizes it as the path tangent.
On real ego trajectories, this reproduces recorded yaw to a mean of \ang{0.4}, with the turning categories no worse than the straight ones, so the synthesis does not distort Epona's maneuver results.
The same pipeline we validate against Vista's external score thus produces Epona's first reported number.

\begin{figure}[htbp]
  \centering
  \input{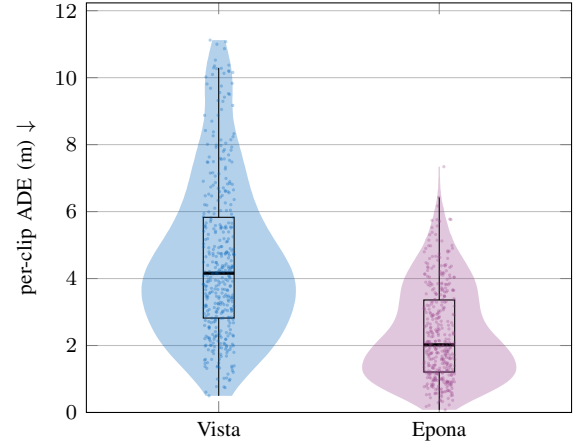}
  \caption{\textbf{Per-clip action-following error (ADE, L1).} Distribution of per-clip ADE (m) for Vista and Epona ($n=400$ each), shown as violin plots with overlaid box plots (median, interquartile range, and whiskers).}
  \label{fig:l1_distribution}
\end{figure}

\begin{figure}[htbp]
  \centering
\colorlet{VistaC}{TUMBlue}
\colorlet{EponaC}{TUMPinkDark}
\begin{tikzpicture}[font=\footnotesize]
\begin{axis}[width=7cm, height=9cm, xlabel={IEC ($\uparrow$)}, ylabel={Category}, xmin=-0.05, xmax=1.05,
  xtick={0,0.5,1},
  ytick={0,1,2,3,4,5,6,7},
  yticklabels={{straight\\accelerating}, stopping, starting, {straight\\decelerating}, {straight const.\\low speed}, {straight const.\\high speed}, {curving\\to right}, {curving\\to left}},
  yticklabel style={align=right}, enlarge y limits=0.12,
  legend pos=south east, ymajorgrids, xmajorgrids]
\addplot+[VistaC, only marks, mark=*, mark size=2pt, error bars/.cd, x dir=both, x explicit]
 table[x=x, y=y, x error plus=ep, x error minus=em] {
x y ep em
0.2800 0.16 0.1367 0.1053
0.1600 1.16 0.1251 0.0766
0.1000 2.16 0.1136 0.0565
0.2400 3.16 0.1341 0.0970
0.9000 4.16 0.0565 0.1136
0.3600 5.16 0.1386 0.1186
0.4800 6.16 0.1349 0.1320
0.1000 7.16 0.1136 0.0565
};
\addlegendentry{Vista}
\addplot+[EponaC, only marks, mark=*, mark size=2pt, error bars/.cd, x dir=both, x explicit]
 table[x=x, y=y, x error plus=ep, x error minus=em] {
x y ep em
0.1200 -0.16 0.1180 0.0638
0.1000 0.84 0.1136 0.0565
0.0800 1.84 0.1084 0.0485
0.2400 2.84 0.1341 0.0970
1.0000 3.84 0.0000 0.0713
0.7600 4.84 0.0970 0.1341
1.0000 5.84 0.0000 0.0713
1.0000 6.84 0.0000 0.0713
};
\addlegendentry{Epona}
\end{axis}
\end{tikzpicture}
  \caption{ \textbf{Action-following consistency by category (IEC, L1).} Per-category instruction-execution consistency (IEC) for Vista and Epona, with \qty{95}{\percent} Wilson confidence intervals (50 clips per category).}
  \label{fig:per_category}
\end{figure}
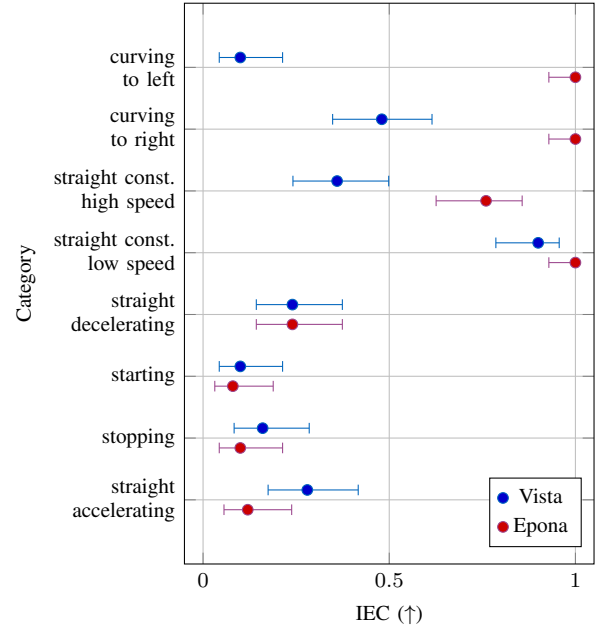

Both models clear L1 (Table~\ref{tab:results_table}): their \gls{iec} of 0.33 (Vista) and 0.54 (Epona) lies well above the $1/8$ chance level, so each follows the commanded maneuver rather than replaying a fixed future.
Epona is the more action-robust on every metric, with a lower \gls{ade} (\qty{2.35}{\meter} against \qty{4.56}{\meter}).
Figure~\ref{fig:l1_distribution} shows the full per-clip distributions.
Epona's shifted lower and tighter and correspondingly has lower \gls{dtw} and higher \gls{sr}.
The per-category breakdown (Figure~\ref{fig:per_category}) locates the difference:
Epona reaches near-perfect \gls{iec} on the curve and constant-speed categories, whereas Vista is weaker and asymmetric between left and right curves.
Both models follow the start and stop transitions poorly, a shared limit at the boundary between rest and motion.


\subsubsection{L2: Action-Following Horizon}
\label{subsec:l2_horizon_bounded_validity}

\begin{figure}[htbp]
  \centering
\colorlet{VistaC}{TUMBlue}
\colorlet{EponaC}{TUMPinkDark}
\begin{tikzpicture}[font=\footnotesize]
\begin{axis}[width=9.2cm, height=6.5cm, xlabel={rollout horizon $h$ (s)},
  ylabel={ADE($h$) (m)}, xmajorgrids, ymajorgrids, legend pos=north west,
  legend cell align=left, legend image post style={line width=1.5pt},
  xmin=0.3, xmax=4.5, ymin=0, ymax=7.5]
\fill[pattern=north east lines, pattern color=TUMGreen!60] (axis cs:0.3,0) rectangle (axis cs:4.5,1.8);
\fill[pattern=north west lines, pattern color=TUMRed!60] (axis cs:0.3,1.8) rectangle (axis cs:4.5,7.5);
\draw[TUMGreenDark, densely dotted, line width=1.2pt] (axis cs:0.3,1.8) -- (axis cs:4.5,1.8) node[pos=0.0, anchor=south west, font=\footnotesize, TUMGreenDark, align=left, fill=white, fill opacity=0.65, text opacity=1, inner sep=1.5pt] {admissible\\($\leq$1.8\,m)};
\addplot[name path=vistahi, draw=none, forget plot] coordinates {(0.400,0.504) (0.800,1.163) (1.200,1.826) (1.600,2.484) (2.000,3.135) (2.400,3.776) (2.800,4.407) (3.200,5.032) (3.600,5.656) (4.000,6.285) (4.400,6.931)};
\addplot[name path=vistalo, draw=none, forget plot] coordinates {(0.400,0.123) (0.800,0.278) (1.200,0.425) (1.600,0.576) (2.000,0.748) (2.400,0.944) (2.800,1.163) (3.200,1.404) (3.600,1.659) (4.000,1.922) (4.400,2.193)};
\addplot[VistaC, opacity=0.18, forget plot] fill between[of=vistahi and vistalo];
\addplot[VistaC, line width=1.5pt, mark=*, mark size=2pt] coordinates {(0.400,0.313) (0.800,0.720) (1.200,1.125) (1.600,1.530) (2.000,1.942) (2.400,2.360) (2.800,2.785) (3.200,3.218) (3.600,3.657) (4.000,4.104) (4.400,4.562)};
\addlegendentry{Vista}
\draw[VistaC, dashed] (axis cs:1.60,0) -- (axis cs:1.60,7.5);
\node[VistaC, anchor=south west, font=\footnotesize] at (axis cs:1.60,0.15) {$h^*=1.6$\,s};
\addplot[name path=eponahi, draw=none, forget plot] coordinates {(0.400,0.265) (0.800,0.590) (1.200,0.924) (1.600,1.263) (2.000,1.607) (2.400,1.956) (2.800,2.310) (3.200,2.664) (3.600,3.019) (4.000,3.376) (4.400,3.744)};
\addplot[name path=eponalo, draw=none, forget plot] coordinates {(0.400,0.052) (0.800,0.128) (1.200,0.209) (1.600,0.286) (2.000,0.354) (2.400,0.422) (2.800,0.502) (3.200,0.598) (3.600,0.707) (4.000,0.824) (4.400,0.951)};
\addplot[EponaC, opacity=0.18, forget plot] fill between[of=eponahi and eponalo];
\addplot[EponaC, line width=1.5pt, mark=*, mark size=2pt] coordinates {(0.400,0.158) (0.800,0.359) (1.200,0.566) (1.600,0.774) (2.000,0.981) (2.400,1.189) (2.800,1.406) (3.200,1.631) (3.600,1.863) (4.000,2.100) (4.400,2.347)};
\addlegendentry{Epona}
\draw[EponaC, dashed] (axis cs:3.20,0) -- (axis cs:3.20,7.5);
\node[EponaC, anchor=south west, font=\footnotesize] at (axis cs:3.20,0.15) {$h^*=3.2$\,s};
\end{axis}
\end{tikzpicture}
  \caption{\textbf{Action-following drift over the rollout horizon (ADE, L2).} Average displacement error (ADE) as a function of rollout horizon $h$ for Vista and Epona. The admissible band marks $\text{ADE}$ at or below \qty{1.8}{\meter}, half a \qty{3.6}{\meter} US lane width~\citep{americanassociationofstatehighwayandtransportationofficialsPolicyGeometricDesign2018} (ego path stays within its own lane); dashed lines indicate the maximum admissible horizon $h^*$, the largest evaluated horizon at which ADE remains within the band.
  Solid curves show the mean ADE over the 400 clips; shaded regions span $\pm1$ standard deviation across clips.}
  \label{fig:horizon_drift}
\end{figure}
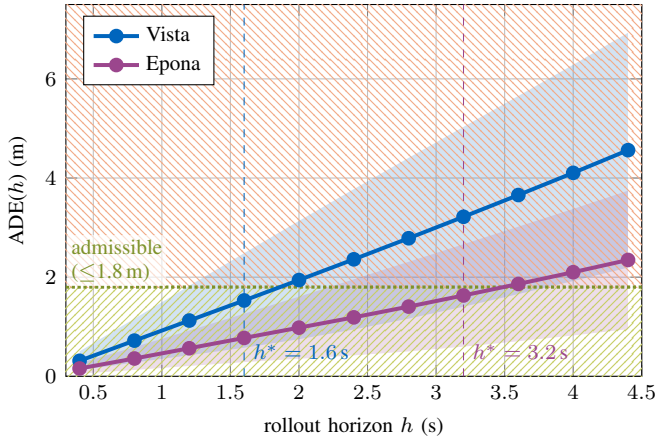

L2 asks whether a model's reactions remain valid over a bounded region of operation, and requires it to declare that envelope rather than be trusted everywhere.
We instantiate neither part of L2's validity core: we neither check reactions against the measured physical dynamics L2 requires as its real-world reference, nor detect and refuse inputs that fall outside the envelope.
What we measure instead is the one component an action-conditioned rollout makes directly observable, the horizon over which action-following stays accurate.
Because it reuses the L1 metric, comparing the recovered trajectory against the commanded one rather than against measured dynamics, it bounds the horizon over which L1 correspondence holds, not validity against reality.
Reusing the L1 rollouts, we truncate each recovered trajectory to its first $h$ seconds, recompute each clip's \gls{ade}, and average over all 400 clips, giving a drift curve \gls{ade}$(h)$ (Figure~\ref{fig:horizon_drift}).
The admissible horizon $h^*$ is the largest evaluated $h$ at which the mean \gls{ade}$(h)$ stays within \qty{1.8}{\meter}, half a \qty{3.6}{\meter} (12~ft) US lane width~\citep{americanassociationofstatehighwayandtransportationofficialsPolicyGeometricDesign2018}, so that the rollout's ego path remains within its own lane.
We read $h^*$ off the sampled horizons without interpolating.
Reporting $h^*$ declares the horizon envelope, and a larger $h^*$ is a longer window over which the model's action-following can be trusted.

Both curves grow roughly linearly but at different rates (Figure~\ref{fig:horizon_drift}, Table~\ref{tab:results_table}).
Vista drifts at about \qty{1.06}{\meter\per\second} and leaves the band at $h^*=\qty{1.6}{\second}$, whereas Epona drifts at about \qty{0.53}{\meter\per\second} and stays admissible until $h^*=\qty{3.2}{\second}$, roughly twice as long.
The ordering does not depend on the threshold: at a more permissive \qty{3.0}{\meter} band Vista reaches \qty{2.8}{\second} while Epona never leaves the band over the measured horizon.
Epona therefore supports action-following claims over a substantially longer window, the envelope within which its L1 verdicts would hold.

\subsection{Decoupling of Generation Quality and Action-Robustness}
\label{subsec:decoupling}

Taken one rung at a time, the preceding measurements each report a single ordering of the two models.
Read together, they reveal a relation that no single rung shows: the ranking by generation quality is the reverse of the ranking by action-robustness.
Vista leads on the L0 visual-fidelity scores (FVD and CD-FVD), whereas Epona leads on every L1 metric and sustains the longer L2 horizon (Table~\ref{tab:results_table}).
Were the two rungs expressions of one underlying quality, the same model would lead on both.
The reversal shows that they are not.
A credibility argument based on visual fidelity would therefore select Vista, while the action-robustness evidence would select Epona.

For a generative \gls{wm} used as a test oracle, this reversal is decisive: a model's L0 visual fidelity does not predict its L1 action-following, the property a closed-loop verdict depends on.
Visual fidelity, therefore, cannot, on its own, accredit a \gls{wm} as a test oracle.
A single model pair does not establish how often the two properties diverge, but one clear divergence is enough to show that they can, and hence that separating the rungs is
necessary rather than redundant.

\subsection{Limitations and Threats to Validity}
\label{subsec:limitations_threats}

Several assumptions enter the measurement and bound the strength of these results.
All L1 and L2 metrics inherit the ACT-Estimator's error: on real clips it recovers trajectories to a mean \gls{ade} of \qty{0.77}{\meter} and classifies maneuvers at about \qty{94}{\percent} accuracy, so displacement differences near that resolution are not meaningful and the reported \gls{iec} carries the estimator's misclassification rate as a noise floor.
The maneuver categories themselves are defined by ACT-Bench's rule-based classifier, several of whose thresholds are underspecified.
The class boundaries are therefore not crisp, which plausibly contributes to the low \gls{iec} of both models on the transition maneuvers.

The comparison is not fully symmetric. Vista is scored on the rollouts released with ACT-Bench, whereas Epona is generated through our adapter.
The scorer is identical and Vista reproduces its published \gls{iec} exactly, but the two models are not conditioned through the same pipeline.
Fidelity is reported at a matched sample size, since \gls{fvd} depends on it.
The values support the within-study comparison rather than the absolute numbers reported elsewhere, and all metrics use a 400-clip stratified subsample rather than the full split.
The two models also differ in architecture, training data, and scale simultaneously, so although their rankings decouple (Section~\ref{subsec:decoupling}), we do not attribute the decoupling to any single cause.

Finally, the instantiation reaches L0, L1, and the horizon component of L2.
These are reachable with compute and an existing instrument.
However, L2 out-of-distribution detection and the failure-attribution and verdict-transfer evidence of L3 and L4 (Section~\ref{sec:Admissibility_Standard}) are not relaxed by computation but require data, methods, or a real-world anchor that the model cannot supply.
The upper bound of the instantiation is therefore evidentiary rather than computational, and falls within L2.

\end{document}